\documentclass[11pt]{article}
\usepackage{acl2014}
\usepackage{times}
\usepackage{url}
\usepackage{subfigure}
\usepackage{latexsym}
\usepackage{enumitem}
\usepackage{graphicx}
\usepackage{array}
\usepackage{booktabs}
\usepackage{amsmath}

\title{Finding Eyewitness Tweets During Crises}

\author{
    Fred Morstatter$^\dagger$, Nichola Lubold$^\dagger$, Heather Pon-Barry$^\dagger$, J{\"u}rgen Pfeffer$^\star$, and Huan Liu$^\dagger$\\
    $^\dagger$Arizona State University\\
    $^\star$Carnegie Mellon University\\
    \{fred.morstatter, nlubold, ponbarry\}@asu.edu, jpfeffer@cs.cmu.edu, huan.liu@asu.edu
}

\date{}

\begin{document}
\maketitle
\begin{abstract}
Disaster response agencies have started to incorporate social media as a source of fast-breaking information to understand the needs of people affected by the many crises that occur around the world. These agencies look for tweets from within the region affected by the crisis to get the latest updates of the status of the affected region. However only 1\% of all tweets are ``geotagged'' with explicit location information. First responders lose valuable information because they cannot assess the origin of many of the tweets they collect. In this work we seek to identify non-geotagged tweets that originate from within the crisis region. Towards this, we address three questions: (1) is there a difference between the language of tweets originating within a crisis region and tweets originating outside the region, (2)  what are the linguistic patterns that can be used to differentiate within-region and outside-region tweets, and (3) for non-geotagged tweets, can we automatically identify those originating within the crisis region in real-time?
\end{abstract}

\section{Introduction}
Every day, users on Twitter post 200 million 140-character messages~\cite{Tsuk13}, called ``tweets'', that often pertain to trending topics in real time. When catastrophic events occur around the world, users flock to the site to post their accounts of the event at an unprecedented scale. As a recent example, the Boston Marathon Bombing was one of the most discussed topics on Twitter in all of 2013.

Because of Twitter's massive popularity and diverse areas of discussion, it has become a tool used by first responders---those who provide first-hand aid---to understand crisis situations and identify the people in the most dire need of assistance~\cite{human-un}. To do this, first responders can survey ``geotagged'' tweets: those where the user has supplied a geographic location. The advantage of geotagged tweets is that first responders know whether a person is tweeting from within the affected region or is tweeting from afar. Tweets from within this region are more likely to contain emerging topics~\cite{Kumar-HT13} and tactical, actionable, information that contribute to situational awareness~\cite{verma2011natural}.

A major limitation of surveying geotagged tweets is that only 1\% of all tweets are geotagged~\cite{morstatter2013sample}. This leaves the first responders unable to tap into the vast majority (99\%) of the tweets they collect. This limitation leads to the question at the heart of this work: can we discover whether a tweet originates from within a crisis region using \emph{only the language used within the tweet}? 

We focus on the language of a tweet as the defining factor of location for three major reasons: (1) the language of Twitter users is dependent on their location~\cite{cheng2010you}, (2) the text is readily available in every tweet, and (3) the text allows for real-time analysis, as opposed to other facets such as friend/follower network data and user history which take a significant amount of time to collect given the limitations of Twitter's APIs. Due to the short time window presented by most crises, first responders need to be able to locate users quickly. By focusing on the tweets' text, we determine whether an individual tweet originates from within the event's location in real-time. 

To identify non-geotagged tweets originating from within a crisis region using only the tweets' language, our approach is to discover structural linguistic patterns that differentiate tweets inside a region from tweets outside the region. Using state-of-the-art Twitter NLP tools, we look at features at the level of words, part of speech (POS) tags, and syntactic constituents. This approach is well-suited for the rapid-response needs of first responders, and it scales to large data sets.  

Existing approaches on predicting tweet location have looked at the problem during non-crisis time. These previous works predict the continuous (latitude/longitude) location of a user given a somewhat substantial history of his tweets. This poses challenges for crisis-time classification: these tweets are hard to get given the API limitations of Twitter, preventing real-time analysis, and not every user tweeting during a time of crisis will have a substantial history to draw upon. In this work, our task is to determine whether a single tweet originates from within the crisis region, a binary classification problem. We examine tweet data from two recent disasters that received a large amount of discussion on social media: the Boston Marathon Bombing and Hurricane Sandy. 

Towards the problem of finding eyewitness tweets during crises, this paper evaluates and verifies three hypotheses. First, we show that linguistic differences exist between the tweets authored inside and outside the affected regions. Second, we show that there are linguistic features that can differentiate tweets originating within a crisis region from those outside. And finally, we show that this classification process can be automated. We build a model that incorporates linguistic features to classify each individual tweet as inside or outside of the crisis region. 
We examine the results of our model including its ability to classify tweets, and identify the most important features for classification during a time of crisis. We find that word unigrams and bigrams and select POS sequences are the most informative. Lastly, we apply our model to non-geotagged tweets and evaluate the added information gleamed from the tweets identified as originating from within the crisis region.

\section{Related Work}
\textbf{Geolocation}, inferring the geographic origin of a particular document, has become a prominent area in the study of social media data. In regards to predicting the origin of a tweet, previous work on geolocation has taken one of two approaches, utilizing either geographical topic models or language-distribution models.  Geographical topic models rely on how the topics users discuss are related to their geographical location; these models identify user preferences and use this information to locate tweets and users.~\newcite{Eisenstein-etal10} first looked at the problem of using latent variables to explain the distribution of text in tweets. This problem was revisted from the perspective of geodesic grids in~\cite{wing2011simple} and further improved by flexible adaptive grids~\cite{roller2012supervised}. Language-distribution models rely on the content of the tweet at a more granular level.~\newcite{cheng2010you} employed an approach that looks at a user's tweets and estimates the user's location based on words consisting of a strong local geographical scope. ~\newcite{han2013stacking} combines a tweet's text with its metadata to predict a user's location. Utilizing either geographical topic models or language-distribution models, a recurring issue is the natural sparsity of a single tweet. Previous works usually handle this by concatenating of all a user's tweets into a single document. In contrast, determining the origin of single tweet is one of the challenges this paper addresses.
  
\textbf{Mass Emergencies} In considering the time-sensitive requirements in responding to crises, first responders have long sought to use Twitter as a source for information during crises. Previous works show that tweets authored during a crisis can hold useful information.~\newcite{de2009omg} studies Twitter’s use as a sensor for crisis information by studying the geographical properties of users tweets. In~\newcite{castillo2011information}, the authors analyze the text and social network of tweets to classify the ``newsworthiness'' of said tweets.~\newcite{Kumar-HT13} uses users who have geotagged their tweets to find emerging topics in crisis data in real time. Investigating linguistic features~\newcite{verma2011natural} shows the efficacy of language features at finding tweets during crisis containing tactical, actionable information, contributing to \textit{situational awareness}. Outside of Twitter,~\newcite{munro2011subword} investigates message prioritization during the 2010 Haiti earthquake through use of subword patterns. Similarly, we find language features and linguistic patterns which can help with the problem proposed here.

\section{Language Differences in Crises}
In order for a language-based approach to be able to distinguish tweets inside of the crisis region, the language used by those in the region during crisis has to be different from those outside. In this section, we verify that there are both regional and temporal differences in the language tweeted before and during a crisis. To start, we introduce the data sets we use throughout the rest of this paper. We then measure the difference in language from various perspectives, seeing that there are definitive and contrasting word distributions. We show that language changes temporally and regionally at the time of the crisis, distinguishing tweets from within the disaster region from those outside.

\subsection{Data}
The Twitter data used in our experiments comes from two different crises: the Boston Marathon bombing and Hurricane Sandy. Both events provoked a significant Twitter response from within and beyond the affected regions. We describe below the mechanisms and methods that were used to collect the dataset and how we partition the data by time and location.

\noindent \textbf{The Boston Marathon Bombing}

On April 15th, 2013 at 2:48 PM Eastern a bombing occurred at the Boston Marathon finish line, hereafter referred to as the ``Boston Bombing''. We collected tweets from before the bombing to several days after the bombing utilizing Twitter's Streaming API\footnote{https://dev.twitter.com/docs/api/1.1/post/statuses/filter}, filtering by geotagged tweets from the continental United States\footnote{We supplied the geographic bounding box to the Streaming API: [-128.6 24.5 -59 50].}. 

\noindent \textbf{Hurricane Sandy}

Hurricane Sandy was a ``superstorm'' that ravaged the Eastern seaboard of the United States during the 2012 hurricane season. Again utilizing Twitter's Streaming API, we collected tweets based on several keywords\footnote{The keywords are as follows: hurricane, sandy, florida, storm, tropical, frankenstorm, sandyde, evacuation, stormde, dctraffic, mdtraffic, vatraffic, baltraffic, nyctraffic, njsandy, nysandy, ctsandy, dcsandy, desandy, njtraffic, shelter, damage, tree, treedown, outage, linedown, power, flood, water, surge, outage, \#hamptons, \#northfork, \#nofo.} pertaining to the storm. Filtering by keywords, this dataset contains both geotagged and non-geotagged data beginning from the day the storm made landfall to several days after it passed. The data was collected using ASU's TweetTracker~\cite{kumar2011tweettracker}.

\begin{table}[t]
    \centering
    \caption{Properties of the Twitter crisis datasets used in this work: Boston Bombing (Boston) and Hurricane Sandy (Sandy).}
    \begin{tabular}{ lcc }
    \toprule
    \textbf{Property} & \textbf{Boston} & \textbf{Sandy} \\
    \midrule
    Start & 09 Apr 00:00 & 29 Oct 20:00 \\
    End & 22 Apr 00:00 & 02 Nov 00:00 \\
    Crisis Start & 15 Apr 14:48 & 29 Oct 20:00 \\
    Crisis End & 16 Apr 00:00 & 30 Oct 01:00 \\
    Epicenter & $42.35$, $-71.08$ & $40.75$, $-73.99$ \\
    Radius & $19$ km & $20$ km \\
    Total Tweets & $24,375,633$ & $200,974$ \\ 
    \bottomrule
    \end{tabular}
    \label{tab:hlprops}
\end{table}

\noindent \textbf{Data Partitioning}

After collecting the datasets, we partitioned the tweets published \emph{during the time of the crises} into two distinct parts based on location:
\begin{enumerate}
\item Inside the region of the crisis (\textbf{IR}).
\item Outside the region of the crisis (\textbf{OR}).
\end{enumerate}

For the Boston Bombing dataset, we are able to extract two additional groups: (1) tweets posted before the time of the crisis (pre-crisis) and inside the region of the crisis (\textbf{PC-IR}) and (2) tweets posted before the time of the crisis (pre-crisis) outside the region of the crisis (\textbf{PC-OR}). 

Due to the disparity in the number of tweets coming from inside the region versus outside the region, we take a time-based sample to obtain \textbf{PC-IR} and \textbf{PC-OR}. We sample the datasets from 10:00 -- 14:48 Eastern on April 9th, 2013. Because the bombing was an abrupt event with no warning, we feel safe in choosing a time period so close to its onset. The number of tweets in each dataset partition is shown in Table~\ref{tab:data_partitions}.

\begin{table}[t]
    \caption{Number of tweets in each partition of the Boston Bombing and Hurricane Sandy datasets.}
    \centerline{\begin{tabular}{lcc}
        \toprule
        \textbf{Group} & \textbf{Boston} & \textbf{Sandy} \\
        \midrule
        \textbf{IR} & 11,601 & 5,017 \\
        \textbf{OR} & 541,581 & 195,957 \\
        \textbf{PC-IR} & 14,052 & N/A \\
        \textbf{PC-OR} & 228,766 & N/A \\
        \bottomrule
    \end{tabular}}
    \label{tab:data_partitions}
\end{table}

\begin{figure}
\centering
\includegraphics[width=0.4\textwidth]{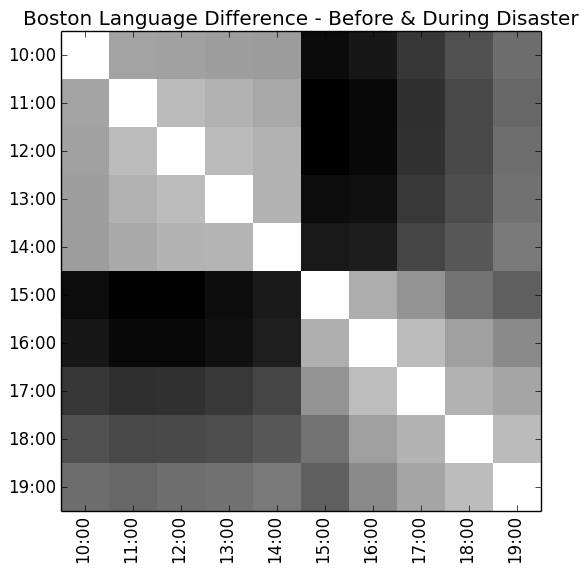}
\caption{Normalized J-S divergence between the text in tweets authored on 2013-04-15, aggregated by hour. Darker cells indicate more divergence between the probability distributions. All hours are presented in the Eastern timezone.}
\label{fig:timediff}
\end{figure}

\subsection{Language Difference Pre-Crisis vs. During Crisis}
Here we investigate whether the language distribution of the tweets changes in the crisis region when a crisis occurs. To perform this analysis, we compare the data generated in the \textbf{PC-IR} and \textbf{IR} partitions of the Boston Bombing datasets. We aggregate the tweets by hour. We compute the difference in the probability distributions for each hour using Jensen-Shannon divergence (J-S)~\cite{lin1991divergence}. Specifically for each pair of hours in the dataset, we compute the J-S divergence of the probability distribution of the words used within those hours. The results of this experiment are shown in Figure~\ref{fig:timediff}. The difference in the hours before the bombing (10:00--14:00) and those after the bombing (15:00--19:00) is visible in the clear contrast between the pre-crisis and during crisis hours. Additionally, we also notice that the tranquil hours are relatively stable.

\begin{figure*}
     \centering
      \subfigure[Tranquil language differences by city. Here we see a relatively low level of divergence.]{
          \label{fig:locdiff_before}
          \includegraphics[width=0.3\textwidth]{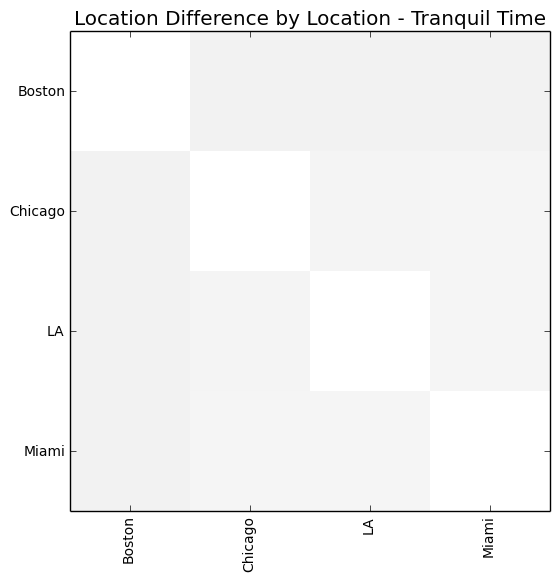}
      }
      \subfigure[Crisis language differences by city during the crisis in the Boston Bombing data.]{%
         \label{fig:locdiff_boston}
         \includegraphics[width=0.3\textwidth]{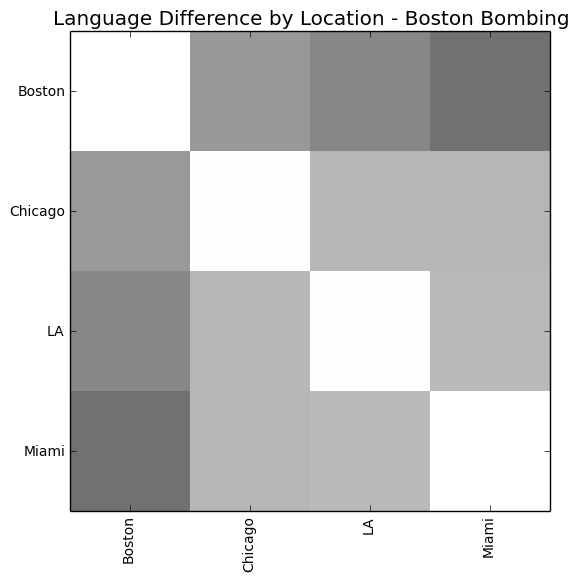}
      }
      \subfigure[Crisis language differences by city during Hurricane Sandy. Here we see a range of values for J-S divergence.]{%
         \label{fig:locdiff_sandy}
         \includegraphics[width=0.3\textwidth]{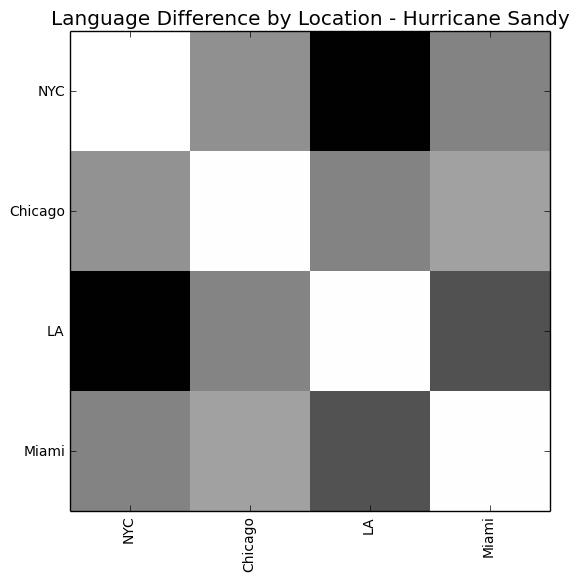}
      }
    \caption{City $\times$ City language distribution comparison. We observe generally similar distributions in the tranquil data. Once a crisis occurs, we see generally higher differences in the language used by location. Darker cells indicate more divergence between the probability distributions. All figures are presented on the same scale.}
   \label{fig:locdiff}
\end{figure*}

\subsection{Language Difference Inside Region vs. Outside Region}
Next we verify that the tweets authored inside of the crisis use different words from those outside the region. We compare the difference in language of four different major US cities: the city affected by the crisis, Chicago, Los Angeles (LA), and Miami. To obtain a baseline, we also compare Boston, Chicago, LA, and Miami during tranquil times using \textbf{PC-IR} and \textbf{PC-OR} datasets. The results of this experiment can be seen in Figure~\ref{fig:locdiff}. The tranquil time comparison, shown in Figure~\ref{fig:locdiff_before}, displays a low divergence between all pairs of cities. In contrast, we see a much wider divergence between the same cities, with Boston and New York as the two cities most affected by the crisis displaying the greatest divergence. 

\section{Methodology}
In this section we outline our methodology for automatically finding inside region \textbf{IR} tweets using linguistic features. We explain our methodology beginning with in-depth discussion of the features, describing how we obtain these features, and ending with the overall model.

\subsection{Linguistic Features}
\label{sec:features}
As Twitter is a conversational, real-time, microblogging site, the structure of tweets offers many opportunities for extracting different types of features that represent the different linguistic properties of informal text. We identify six linguistic feature classes at varying linguistic levels.  At the \emph{word-based} level, we look at unigrams and bigrams. To examine the effect of linguistic structure at the \emph{part of speech} level, we extract POS n-grams where $n = 1, 2, 3$ utilizing two different, unique tag sets. At the  level of \emph{syntactic constituents}, we also extract shallow parse tag n-grams where $n = 1, 2, 3$. Finally, combining all levels of linguistic analysis, we create a mixed-class composed of \emph{word-based}, \emph{part of speech}, and \emph{syntactic constituent} features which are sensitive to crisis, and label this class ``crisis sensitive." Organizing our features this way allows us to analyze and incorporate the affect of different linguistic levels in regards to our problem.  



\noindent \textbf{Unigrams} \& \textbf{Bigrams}

We extract the raw frequency counts of the unigrams and bigrams. Unigrams are commonly selected as features for geo-location \cite{roller2012supervised,Eisenstein-etal10,hong2012discovering}.  
  

\noindent \textbf{ARK Tag Set}

The ARK tag set was developed specifically for POS tagging in the dynamic and informal world of tweets. As a result, the tag set is coarser than the Penn-Treebank (PTB) tag set. For example, there is a single verb tag ``V'' in the ARK data set compared to six verb tags found in the Penn-Treebank tag set. In addition, the ARK tag set introduces tags specific to Twitter such as ``\#'' to capture Twitter hashtags. The advantage is that particular tag classes are less sparse than with the Penn-Treebank tag set, and we can identify Twitter-specific entities in the data.

\noindent \textbf{PTB Style Tag Set}

We also extract the Penn Treebank style part-of-speech tags for each of the words in the tweet. Because the PTB tag set is more fine-grained, it provides a nice contrast and comparison to the ARK tag set. By comparing both sets of tags as feature classes, we can measure the effectiveness of the fine-grained versus coarse-grained applications provided by both sets.


\noindent \textbf{Shallow Parsing Tags}

In addition to the POS tags, we also extract shallow parsing tags along with the headword associated with the tag. For example, in the noun phrase ``the movie'' we would extract the headword ``movie'' and represent it as [...movie...]$_{NP}$. Shallow parsing identifies the constituents of a sentence, breaking down the sentence into grammatical components or phrases. The underlying motivation with this class is that the shallow parsing groups and associated headwords can give more insight into the syntactic differences of \textbf{IR} tweets versus \textbf{OR} tweets. 

\noindent \textbf{Crisis-Sensitive (CS) Features}

\textbf{Part-of-Speech Patterns} To reduce the number of dimensions in the dataset and possible noise, specific POS tag patterns are extracted in both their ARK and PTB forms; we extract these patterns both as the stand-alone tag sequence and as the word/tag sequence by including the content which was tagged with the POS pattern. To find the patterns, we analyze the Boston Marathon data set, reviewing the raw frequencies of POS tag sequences from tweets originating from within the region. We find sequences which are used more widely during the time of this disaster. The extracted ARK patterns are as follows: N, A, !, N R, L A, N P, P D N, L A !, A N P. The motivation behind these patterns rises from ~\newcite{munro2011subword}, who finds that particular sub-word patterns improved the accuracy of identifying actionable text in SMS messages coming from the 2010 Haiti Earthquake. We apply these features to the Hurricane Sandy data set to validate whether the features are generalizable to crisis and discuss this in the results. 


\textbf{Prepositional Phrase Pattern} Within the crisis sensitive class, we incorporate several linguistic levels; in addition to the part of speech level addressed above, a combined pattern incorporating the preposition \textit{in} and its associated noun targets both part of speech and syntactic constituents. Specifically we extract any patterns reflecting the form [in ... /N]$_{PP}$.  An example would be ``in Boston.''  The motivation is that generally ``in" as a preposition is used to indicate a location, nonspecific times, shapes, color, or to indicate a belief. While not all of these will be pertinent in a crisis, the hypothesis is that the frequency of this feature will increase during a crisis due to the part of speech types it normally introduces. 

\textbf{Existential \textit{there}} The final feature included under the crisis sensitive class involves extracting verbs in relationship to the existential \textit{there}.  \newcite{lakoff1987women} defines the existential \textit{there} as ``a mental space in which a conceptual entity is to be located''. In contrast to the deictic ``there,'' the existential \textit{there} does not refer to a specific location and is not a locative adverb. It is usually the grammatical subject and describes an abstraction. An example of the different usages of ``there'' can be seen below:
\begin{quote}
\textrm{``\textit{There} is a bomb" (existential)} \newline
\textrm{``I saw a man over \textit{there}." (adverbial)}
\end{quote}
Two interpretations of ``there'' are: (1) it is derived from a locative adverbial or (2) is a noun phrase inserted for syntactic and pragmatic reasons \cite{breivik1981interpretation}. Regardless of which interpretation is taken, the use of the existential \textit{there} may be more likely to be employed when uncertain conditions prevail, such as in a mass emergency. The hypothesis is that the frequency of the existential ``there'' in conjunction with the succeeding verb may be indicative of a tweet's origin within the disaster region.   

\subsection{Data Preprocessing}
To obtain the features we describe in detail in \ref{sec:features}, we perform the following preprocessing techniques for each tweet:
\begin{itemize}
    \item Tokenization: We tokenize each tweet utilizing a tokenizer built explicitly for Twitter data and provided by~\newcite{owoputi2013improved}. This process also standardizes word case.
    \item POS Tagging: We perform part of speech (POS) tagging  on each tweet. The default settings for the ARK tagger \cite{owoputi2013improved} output POS tags in a special ``ARK'' tagset, designed specifically for Twitter. In parallel, we extract the Penn-Treebank style tags from the tweets using the ARK tagger configured with the model proposed by~\newcite{derczynski2013twitter}.
    \item Shallow Parse Tagging: We use~\newcite{Ritter11namedentity} to perform shallow parsing. This process identifies the syntactic constituents in the tweets, such as noun phrases and verb phrases.
\end{itemize}

\subsection{Automatically Finding In-Region Tweets}
\label{sec:classifier}
To classify tweets as \textbf{IR} or \textbf{OR} automatically, we take a machine learning approach, employing a Na{\"i}ve Bayes classifier to make predictions about a tweet's location. We select this classification algorithm for its simplicity and its success in other processing tasks involving natural language. As input to the classifier, we vectorize the tweet by extracting the features from the tweet's text. Each of our features are represented as raw frequency counts of the number of times they occur within the tweet. The model then outputs its prediction of whether the tweet is inside region (\textbf{IR}) or outside region (\textbf{OR}).

\section{Experiments}
Here, we assess the effectiveness of our linguistic features at the task of identifying tweets originating from within the crisis region. To do this we use the classifier described in Section~\ref{sec:classifier} configured with different sets of feature classes. In doing this we tackle our final two questions simultaneously: we find the features that can differentiate the two classes of users, and we show that this process can indeed be automated.

\subsection{Experiment Procedure}
\label{sec:expsetup}

Before running each experiment, we partition the data to ensure a 50/50 split of \textbf{IR} and \textbf{OR} instances. Using the classifier described in Section~\ref{sec:classifier}, we perform $3\times5$-fold cross validation on the data. This process splits the data into 5 folds and iteratively trains the classifier on 4 of those folds (80\% of the data), and tests on the fifth fold (the remaining 20\%). This process is repeated 3 times, shuffling the rows each time. After this process is completed, we obtain the results for the accuracy, precision, and recall of the classifier in the form of 15 readings. The average of these 15 readings is reported in the results.

We compare with a ``select-all'' baseline. This baseline classifier labels all tweets as \textbf{IR}. Because of the nature in which we construct our data for each experiment, the baseline will have an accuracy of 50\%, a precision of 50\%, and a recall of 100\%. All precision and recall values reported are from the perspective of the \textbf{IR} class.

Where appropriate, we compare with another baseline: the model proposed in~\newcite{roller2012supervised}. We adapt this model to our problem by mapping its continuous latitude/longitude output to our binary view of the world. We use the ``Epicenter'' and ``Radius'' values shown in Table~\ref{tab:hlprops} to determine if the latitude/longitude point determined by the classifier falls within our \textbf{IR} class.

\subsection{Individual Feature Class Analysis}
To analyze how each feature class contributes to the task, we test each class individually. Table~\ref{tab:individual_features} shows the results. We find that word unigrams and bigrams obtain the highest scores in both crises. We also notice that our crisis-sensitive features do not perform in either of the crises by themselves. 

\begin{table*}[t]
    \centering
    \caption{Individual feature class performance compared to the top three performing feature class combinations from each data set. For the top performing combinations, ``CS'' refers to the Crisis Sensitive feature class; ``---'' indicates that this was not a top performing combination for that data set. }
    \begin{tabular}{|l|l|l|l||l|l|l|}
    \hline   & \multicolumn{3}{ c|| }{\textbf{Boston Bombing}} & \multicolumn{3}{ c| }{\textbf{Hurricane Sandy}} \\
    \hline \textbf{Individual Feature Performance} & \textbf{Prec.} & \textbf{Recall} & \textbf{F1} & \textbf{Prec.} & \textbf{Recall} & \textbf{F1} \\ \hline
     Unigram & 0.843 & 0.803 & \textbf{0.822} & 0.905 & 0.772 & 0.834 \\
     Bigram  & 0.770 & 0.752 & 0.761 & 0.910 & 0.797 & \textbf{0.849} \\
     ARK POS & 0.605 & 0.614 & 0.609 & 0.745 & 0.678 & 0.710 \\
     PTB POS & 0.651 & 0.643 & 0.647 & 0.828 & 0.723 & 0.772 \\
     Shallow Parse & 0.722 & 0.686 & 0.704 & 0.750 & 0.682 & 0.714 \\
     Crisis-Sensitive (CS) & 0.651 & 0.605 & 0.627 & 0.696 & 0.690 & 0.693 \\
    \hline 
    \hline \textbf{Top Feature Combinations} & \textbf{Prec.} & \textbf{Recall} & \textbf{F1} & \textbf{Prec.} & \textbf{Recall} & \textbf{F1} \\ \hline

    Unigram + Bigram & 0.853 & 0.805 & 0.828 & 0.942 & 0.820 & 0.877\\
    Unigram + Bigram + Shallow Parse & 0.892 & 0.771 & 0.828 & --- & --- & --- \\
     Unigram + Bigram + Shallow Parse + CS & --- & --- & --- & 0.956 & 0.803 & 0.873 \\
     \textbf{Unigram + Bigram + CS} & 0.857 & 0.806 & \textbf{0.831} &  0.947 & 0.826 & \textbf{0.882} \\
     All Features & 0.897 & 0.742 & 0.812 & 0.960 & 0.786 & 0.864\\
    \hline
    \hline
    \end{tabular}
    \label{tab:individual_features}
\end{table*}

\subsection{Top Performing Feature Class Combinations}
\label{sec:topcombos}
While looking at the individual feature classes provides an understanding of each individual class's contribution to the problem, individual feature classes may combine for results that are superior to any one class alone. We find the top feature class combinations by assessing the F1-score of all $\sum_{i=1}^{6} \binom{6}{i} = 63$ possible combinations. We report the top three in Table~\ref{tab:individual_features}.

We see that in both crises all of the top performing feature combinations contain both the bigram and unigram feature classes. The breadth of previous work on geo-location models with Twitter data has suggested that unigrams have had unchallenged success as a feature class. Our top performing feature combinations demonstrate that bigrams in combination with unigrams have added utility.  In addition, we see that while as an individual feature class, the crisis-sensitive features did not stand out, CS is present in the Boston data set as a contributing class in the top performing combination (based on F1-score), and CS is also present in the Hurricane Sandy set in both the first and third highest performing combinations, suggesting that CS does have value when considered with the word-based classes. This feature class was trained on Boston Bombing data, so its presence in the top groups from Hurricane Sandy is an indication these features are general, and may be useful for finding users in these and future crises.

\subsection{Performance Under Class Imbalance}
In our experimental setup, we guarantee that the classifier will have a 50/50 split of positive/negative training examples. In the real world this is rarely the case. In fact, we find a major class imbalance in both of our two real-world datasets: $\frac{|IR|}{|IR| + |OR|} = 0.021$ for the Boston Bombing and $0.025$ for Hurricane Sandy.

To understand how our feature classes perform under differing levels of class imbalance, we repeat the experiments from the previous two sections, varying the class imbalance from all-positive (all-\textbf{IR}) to all-negative (all-\textbf{OR}) instances. We calculate the receiver operating characteristic (ROC) area under the curve (AUC) score for each individual feature class and for each of the top-performing combinations of feature classes. These results are shown in Table~\ref{tab:individual_feature_set_auc}. Our combinations outperform our baseline of the~\newcite{roller2012supervised} model by 9.2\% in the Boston Bombing and 31.8\%. One interesting result is that while combinations seem ideal in the case of the 50/50 split, the bigram feature set achieves the best results in the case of Hurricane Sandy.

\begin{table*}[t]
    \centering
    \caption{ROC AUC scores for the individual feature classes and top performing feature class combinations trained with differing levels of class imbalance. Comparison to~\newcite{roller2012supervised}.}
    \begin{tabular}{| p{6cm} | p{3cm} | p{3cm} |}
    \hline \textbf{Individual Feature Classes} & \textbf{Boston Bombing} & \textbf{Hurricane Sandy} \\
    \hline Unigram & \textbf{0.874} & 0.833 \\
    \hline Bigram  & 0.841 & \textbf{0.900} \\
    \hline ARK POS & 0.674 & 0.725 \\
    \hline PTB POS & 0.718 & 0.725 \\
    \hline Shallow Parse & 0.736 & 0.720 \\
    \hline Crisis-Sensitive (CS) & 0.683 & 0.812 \\
    \hline 
    \hline \textbf{Top Combinations} & \textbf{Boston Bombing} & \textbf{Hurricane Sandy}\\
    \hline Unigram + Bigram & \textbf{0.884} & 0.891 \\ 
    \hline Unigram + Bigram + CS & \textbf{0.884} & \textbf{0.894} \\
    \hline Unigram + Bigram + Shallow & \textbf{0.884} & --- \\
    \hline Unigram + Bigram + CS + Shallow & --- & 0.855 \\
    \hline
    \hline \textbf{\newcite{roller2012supervised}} & 0.792 & 0.582 \\
    \hline 
    \hline
    \end{tabular}
    \label{tab:individual_feature_set_auc}
\end{table*}

\section{Analysis}

\subsection{Top Performing Individual Features}
In this section we zoom in to see which individual features within the classes give the best performance. To perform this analysis we make a modification to the experiment setup described in Section~\ref{sec:expsetup}: we replace the Na{\"i}ve Bayes classifier with a Logistic Regression classifier. The benefit of using Logistic Regression for this task is that the coefficients it uses for each individual feature provide a weight that can be used to assess that feature's importance. We report the top three features from each set in Table~\ref{tab:individual_sub_features}. 

\begin{table*}[t]
    \centering
    \caption{Top 3 individual features within each feature class for both data sets; these results highlight how the features which are highly indicative of a tweet originating from within the crisis region contain suggestions of how crisis information is disseminated. ``CS'' refers to the Crisis Sensitive feature class.} 
    \begin{tabular}{|l|p{6.25cm}|p{6.25cm}|}
    \hline \textbf{Feature Class} & \textbf{Boston Bombing} & \textbf{Hurricane Sandy} \\
    \hline Unigram & \#prayforboston, boston, explosion & @kiirkobangz, upset, staying \\
    \hline Bigram & in boston, the marathon, i'm safe & railroad :, evacuation zone, storm warning \\
    \hline ARK POS & ``P \$ \^{}~'', ``L !'', ``! R P'' & ``P \#'', ``$\sim$ \^{} A'', ``@~@~\#'' \\
    \hline PTB POS& CD NN JJ, CD VBD, JJS NN TO & USR DT JJS, VB TO RB, IN RB JJ\\
    \hline Shallow Parse & [...explosion...]$_{NP}$, [...marathon...]$_{NP}$, [...bombs...]$_{NP}$ & [...bomb...]$_{NP}$, [...waz...]$_{VP}$, [...evacuation...]$_{NP}$ \\
    \hline Crisis-Sensitive & [in boston/N]$_{PP}$, [for boston/N]$_{PP}$, i'm/L safe/A & while/P a/D hurricane/N, [in http://t.co/UxkKJLoX/N]$_{PP}$, of/P my/D house/N \\
    \hline
    \hline 
    \end{tabular}
    \label{tab:individual_sub_features}
\end{table*}

When examining these results, the individual unigram and bigram features with the most weight have a clear semantic relationship to the crisis, which makes sense given the success of both of these classes as indicators for the origin of a tweet. Another observation is that between the two events, the top features in Hurricane Sandy are more concerned with user-user communication. In fact the most useful unigram in classification was the user mention ``@kiirkobangz''. The ARK POS trigram ``@~@~\#'' indicates that users are trying to spread information between each other. This focus on communication could be as a result of the warning that came from the storm.   

\begin{figure*}[t]
     \begin{center}
        \subfigure[Word bigrams from inside-region (\textbf{IR}) geotagged tweets.]{
            \label{fig:justgeo}
            \includegraphics[width=0.5\textwidth]{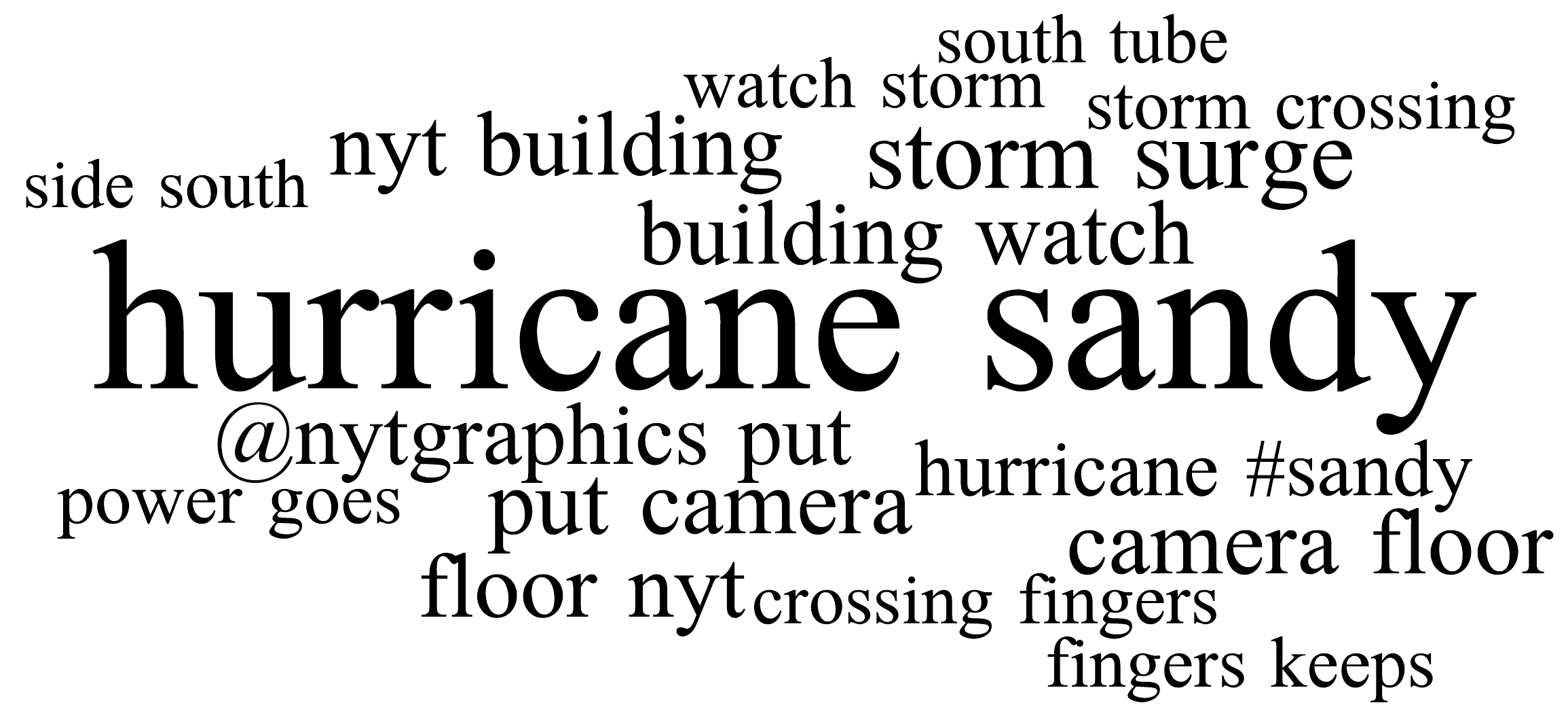}
        }~
        \subfigure[Word bigrams from inside-region (\textbf{IR}) geotagged tweets and tweets classified as IR by our model.]{%
           \label{fig:withnewtweets}
           \includegraphics[width=0.5\textwidth]{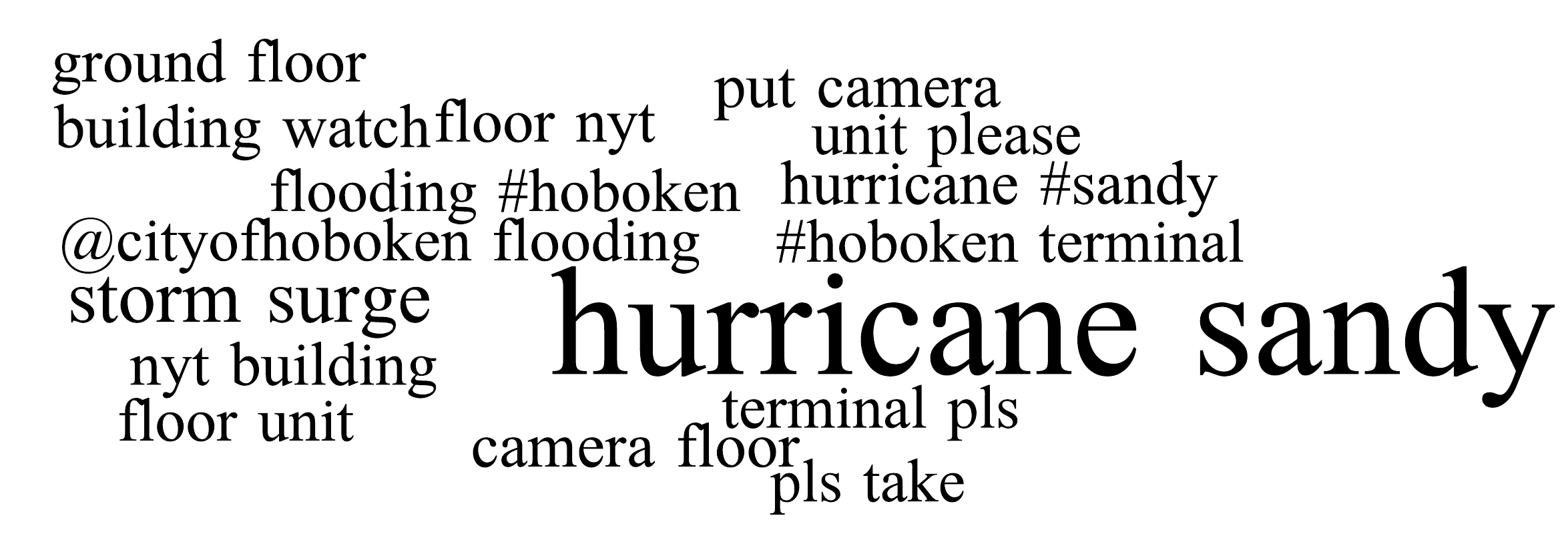}
        }
    \end{center}
    \caption{Tag clouds of top words from the original \textbf{IR} set from the first two hours of Hurricane Sandy.}
   \label{fig:casestudy}
\end{figure*}

\subsection{Use Case: Model in Action}
In the previous experiments we assessed the ability of our model to find tweets in the crisis region. In this section we apply our model to a real crisis dataset to see the real-world benefit. We collected the tweets for our Hurricane Sandy dataset by matching certain keywords within the text. This method of data collection yields both geotagged and non-geotagged tweets. We train our model using the geotagged tweets from the first two hours after Hurricane Sandy made landfall. We use the trained model to classify the non-geotagged tweets that were produced during this time.

During the first two hours after the storm made landfall, we find 2,313 tweets that originate from within the crisis region. Our model finds an additional 739 tweets that originate within the crisis region. To understand the impact made by our model, we analyze just the geotagged tweets, and see the difference when we combine the tweets selected by the model. Figure~\ref{fig:casestudy} shows word bigram clouds from both of these configurations. In the cloud for just the geotagged tweets shown in Figure~\ref{fig:justgeo}, we see general discussion about the storm along with tweets pertaining to the storm, the power outage, and the flooded New York Times building. When we add the tweets discovered by the model, we see many tweets discussing flooding in the city of Hoboken, New Jersey\footnote{The city of Hoboken falls within our crisis radius, so we expect to see flooding-related  tweets in our \textbf{IR} group.}. This is information that is apparent in the new dataset, but was previously not emphasized in just the \textbf{IR} data. This is actionable information that can be used by first responders to direct relief to this area.

\section{Conclusion and Future Work}
This paper addresses the challenge of finding tweets that originate from a region that is affected by a crisis. We make this distinction using only the language of the tweet. We verify three hypotheses that each contribute to answering this question: (1) whether the tweets authored from a crisis region differ in terms of their language from tweets authored outside the crisis region, (2) whether there are features that help distinguish tweets from the crisis region, and (3) whether we can automate this task. We find that the tweets authored from within the crisis region do differ, from both tweets published during tranquil time periods and from tweets published from other geographic regions. Applying these differences, we identify six feature classes that may help to distinguish tweets authored during crisis, in the region. We build a classifier based on these features to automate the process of identifying the in region tweets, finding that our classifier performs well and that this approach is suitable for attacking this problem. 

Overall, our findings indicate that unigrams and bigrams may be the most useful features in answering this question. Upon inspecting the performance of each individual feature class, we see that our crisis-sensitive feature class is one of the lowest performing in both disasters. However, we also find that our crisis-sensitive feature class performs very well when combined with unigrams and bigrams, contributing to the highest performing feature class combination in both the Boston Bombing and the Hurricane Sandy data sets. This implies that there are a subset of structural and part-of-speech features that can add context to the words used in a crisis to give a clearer picture of a tweet's location. The breadth of previous work on geolocation models with Twitter data has suggested that unigrams have had unchallenged success, but our results demonstrate that unigrams in combination with other feature classes such as bigrams and the crisis-sensitive features have potential for stronger classification performance.

Future work includes incorporating the wealth of tweets published before the disaster occurs to make more accurate predictions. Future work may also consider additional features. Given the changing mood of Twitter which can be seen to shift with time~\cite{golder2011diurnal}, we hypothesize that sentiment may prove a useful feature for our task. Future work also seeks to find well-tuned classifiers that can increase the accuracy found in this work. 

\section*{Acknowledgments}
This work is sponsored in part by the Office of Naval Research, grants N000141010091 and N000141110527, and the Ira A. Fulton Schools of Engineering Dean's Office, through fellowships to F. Morstatter and N. Lubold. The authors thank Alan Ritter, Stephen Roller, and the ARK research group at CMU for sharing their tools and providing helpful suggestions and assistance throughout this work.

\bibliographystyle{acl}
\bibliography{references}

\end{document}